\documentclass{article}
\usepackage{spconf,amsmath,graphicx,amsfonts}
\usepackage{xcolor}
\usepackage{algorithm}
\usepackage{algpseudocode}

\usepackage{tikz,pgfplots,pgfplotstable}   
\usepgfplotslibrary{groupplots,fillbetween,colorbrewer,statistics}
\usetikzlibrary{intersections,calc,arrows,matrix,spy,pgfplots.statistics, pgfplots.colorbrewer}
\pgfplotsset{plot coordinates/math parser=false}
\usepackage{color}
\definecolor{magenta}{rgb}{0.7,0,0.7}
\definecolor{darkred}{rgb}{0.7,0,0}
\definecolor{darkgreen}{rgb}{0,0.5,0}
\definecolor{darkblue}{rgb}{0,0,0.7}
\definecolor{SkyBlue}{rgb}{0.53, 0.81, 0.92}

\usepackage[colorlinks]{hyperref}

\title{UVTomo-GAN: An adversarial learning based approach for unknown view X-ray tomographic reconstruction}
\name{Mona Zehni,  Zhizhen Zhao}
\address{Department of ECE and CSL, University of Illinois at Urbana-Champaign}
\begin{document}
%
\maketitle
%
\begin{abstract}
Tomographic reconstruction recovers an unknown image given its projections from different angles. State-of-the-art methods addressing this problem assume the angles associated with the projections are known a-priori. Given this knowledge, the reconstruction process is straightforward as it can be formulated as a convex problem. Here, we tackle a more challenging setting: 1) the projection angles are unknown, 2) they are drawn from an unknown probability distribution. In this set-up our goal is to recover the image and the projection angle distribution using an unsupervised adversarial learning approach. For this purpose, we formulate the problem as a distribution matching between the real projection lines and the generated ones from the estimated image and projection distribution. This is then solved by reaching the equilibrium in a min-max game between a generator and a discriminator. Our novel contribution is to recover the unknown projection distribution and the image simultaneously using adversarial learning. To accommodate this, we use Gumbel-softmax approximation of samples from categorical distribution to approximate the generator's loss as a function of the unknown image and the projection distribution. Our approach can be generalized to different inverse problems. Our simulation results reveal the ability of our method in successfully recovering the image and the projection distribution in various settings. 
\end{abstract}
\vspace{-5pt}
\vspace{-5pt}
\section{Introduction}
\label{sec:intro}
\vspace{-7pt}
X-ray computed tomography (CT) is a popular imaging technique that allows for non-invasive examination of patients in medical/clinical settings. In a CT setup, the measurements, i.e. projections, are modeled as the line integrals of the underlying 2D object along different angles. The ultimate goal in CT reconstruction is to recover the 2D object given a large set of noisy projections.

If the projection angles are known, the tomographic reconstruction problem is often solved via Filtered Back-projection (FBP), direct Fourier methods~\cite{start1981} or formulated as a regularized optimization problem~\cite{Gong2019}. However, the knowledge of the projection angles is not always available or it might be erroneous, which adversely affects the quality of the reconstruction. To account for the uncertainty in the projection angles, iterative methods that solve for the 2D image and the projection angles in alternating steps are proposed in~\cite{Cheikh2017}. While proven effective, these methods are computationally expensive and sensitive to initialization.


Recently, the use of deep learning (DL) approaches for tomographic reconstruction has surged. DL-based CT reconstruction methods in sparse-view regimes learn either a mapping from the sinograms to the image domain~\cite{Bo2018,Ge2020an} or a denoiser that reduces the artifacts from the initial FBP reconstructed image from the sinogram~\cite{jin2017, Quan2018,chen2017LDCT,han2018framelet,kang2018dlframe,yang2018WGAN}. Furthermore, DL-based sinogram denoising or completion is proposed in~\cite{dong2019, li2019inpaint}. Solving the optimization formulation of tomographic reconstruction along the gradient descent updates with machine learning components is suggested in~\cite{Adler2017, Adler2018}. While these methods rely on the knowledge of the projection angles, they also require large paired training sets to learn from. However, here we address a more challenging problem where the projection angles are unknown in advance. 

 \begin{figure}
     \centering
     \includegraphics[width=1 \linewidth]{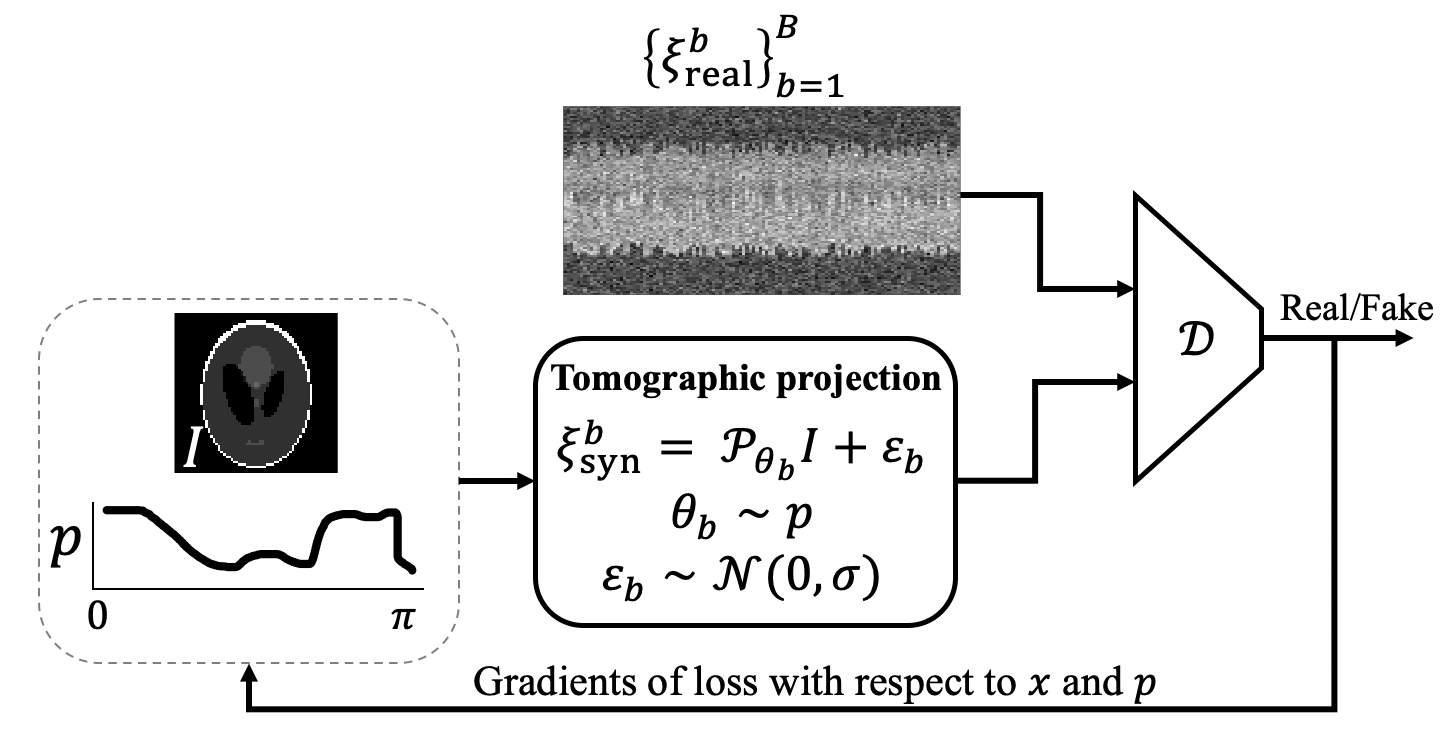}
     \vspace{-20pt}
     \caption{An illustration of our pipeline.}
     \vspace{-18pt}
     \label{fig:pipeline}
 \end{figure}

To overcome the challenges for unknown view CT reconstruction, we propose UVTomo-GAN, an unsupervised adversarial learning based approach for tomographic reconstruction with unknown projection angles.
Our method is unsupervised, thus there is no need for large paired training sets. Our approach benefits from the proven potential of generative adversarial networks (GAN)~\cite{NIPS2014_5423} to recover the image and projection angle distribution that match the given projection measurements in a distribution sense.  Our approach is mainly inspired by CryoGAN~\cite{cryogan}. Unlike CryoGAN, we have a more challenging setting, as we assume that the distribution of the projection angles is \textit{unknown}. Therefore, we seek to recover this distribution alongside the image.
We show that the original generator's loss involves sampling from the projection angles distribution which is non-differentiable. To allow for back-propagation through this non-differentiable operator, we alter the training loss at the generator side using gumbel-softmax approximation of samples from a categorical distribution~\cite{gumbelsoftmax}. 
Our proposed idea is general and can be applied to a wide range of inverse problems with similar setups. Our results confirm the potential of our method in unknown view tomographic reconstruction task under different noise regimes. Our implementation is available at \url{https://github.com/MonaZI/UVTomogan}.
\vspace{-5pt}

\vspace{-0.21cm}
\section{Projection Formation Model}
\label{sec:model}
\vspace{-7pt}
We assume the projection formation model for X-ray CT as,
\vspace{-5pt}
\begin{equation}
    \xi_{\ell} = \mathcal{P}_{\theta_\ell} I + \varepsilon_\ell, \, \ell \in \{1, 2, ..., L\}
    \vspace{-5pt}
    \label{eq:proj_noisy}
\end{equation}
where $I: \mathbb{R}^2 \rightarrow \mathbb{R}$ is an unknown 2D compactly supported image we wish to estimate. $\mathcal{P}_{\theta_\ell}$ denotes the tomographic projection operator that takes the line integral along the direction specified by $\theta_\ell \in [0, \pi]$, i.e.
\vspace{-8pt}
\begin{equation}
    (\mathcal{P}_{\theta_\ell} I) (x) = \int\limits_{-\infty}^{\infty} I(R^T_{\theta_\ell} \mathbf{x}) dy \vspace{-10pt}
\end{equation}
where $\mathbf{x} = [x, y]^T$ represents the 2D Cartesian coordinates and $R_{\theta_{\ell}}$ is the 2D rotation matrix specified by angle $\theta_\ell$. Here, we assume that $\{\theta_{\ell}\}_{\ell=1}^L$ are unknown and are randomly drawn from an \textit{unknown} distribution $p$. Finally, the discretized projections are contaminated by additive white Gaussian noise $\varepsilon_\ell$ with zero mean and variance $\sigma^2$. An unbiased estimator of $\sigma$ can be obtained from the variance of the projection lines but here we assume that $\sigma$ is known.

In this paper, our goal is to recover the underlying image $I$ and the unknown distribution of the projection angles $p$, given a large set of noisy projection lines, i.e. $\{\xi_{\ell}\}_{\ell=1}^{L}$.
\vspace{-5pt}

\vspace{-0.22cm}
\section{Method}
\label{sec:method}
\vspace{-7pt}
Our approach involves recovering $I$ and $p$ such that the distribution of the projection lines generated from $I$ and $p$ matches the distribution of the real projection lines. To this end, we adopt an adversarial learning framework, illustrated in Fig.~\ref{fig:pipeline}. 

Our adversarial learning approach consists of a discriminator $\mathcal{D}_{\phi}$ and a generator $\mathcal{G}$. Unlike classic GAN models, we replace the generator $\mathcal{G}$ by the a-priori known forward model defined in~\eqref{eq:proj_noisy}. The generator's goal is to output projection lines that match the distribution of the real projection dataset $\{\xi_{\textrm{real}}^{\ell}\}_{\ell=1}^L$ and fool the discriminator. For our model, the unknowns we seek to estimate at the generator side are the image $I$ and the projection angle distribution $p$. On the other hand, the discriminator $\mathcal{D}_{\phi}$, parameterized by $\phi$, tries to distinguish between real and fake projections.

Similar to~\cite{cryogan}, we choose Wasserstein GAN~\cite{wgan} with gradient penalty (WGAN-GP)~\cite{wgangp}. Our loss function and the mini-max objective for $I$, $p$ and $\phi$ are defined as,
\vspace{-5pt}
{\small
\begin{align}
    &\mathcal{L}(I, p, \phi) \! = \! \sum\limits_{b=1}^{B} \! \mathcal{D}_{\phi} (\xi^b_{\textrm{real}}) \! - \! \mathcal{D}_{\phi} (\xi^b_{\textrm{syn}}) \!  + \! \lambda \! \left( \Vert \nabla_{\xi} \mathcal{D}_{\phi}(\xi^b_{\textrm{int}}) \Vert \! - \! 1 \! \right)^2  \label{eq:loss_function}\\
    &\widehat{I}, \widehat{p} = \arg \min_{I, p} \max_{\phi} \mathcal{L}(I, p, \phi), \label{eq:minmax}
\end{align}}%
where $\mathcal{L}$ denotes the loss as a function of $I$, $p$ and $\phi$. $B$ and $b$ denote the batch size and the index of a sample in the batch respectively. Also, $\xi_{\textrm{real}}$ mark the real projections while $\xi_{\textrm{syn}}$ are the synthetic projections from the estimated image $\widehat{I}$ and projection distribution $\widehat{p}$ with $\xi_{\textrm{syn}}\! =\! \mathcal{P}_{\theta} \widehat{I} \!+ \!\varepsilon$, $\theta \!\sim\! \widehat{p}$ and $\varepsilon \! \sim \! \mathcal{N}(0, \sigma)$. Note that the last term in~\eqref{eq:loss_function} is the gradient penalty with weight $\lambda$ and roots from the Liptschitz continuity constraint in a WGAN setup. We use $\xi_{\textrm{int}}$ to denote a linearly interpolated sample between a real and a synthetic projection line, i.e. $\xi_{\textrm{int}}\! = \!\alpha \, \xi_{\textrm{real}}\! +\! (1\!-\! \alpha) \, \xi_{\textrm{sim}}, \, \alpha \! \sim \! \textrm{Unif}(0, 1)$.
Note that~\eqref{eq:minmax} is a min-max problem. We optimize~\eqref{eq:minmax} by alternating updates between $\phi$ and the generator's variables, i.e. $I$ and $p$, based on the associated gradients.

Given $\mathcal{D}_{\phi}$, the loss that is optimized at the generator is,
\vspace{-5pt}
\begin{equation}
    \mathcal{L}_{\mathcal{G}}(I, p) = - \sum\limits_{b=1}^{B} \mathcal{D}_{\phi} (\mathcal{P}_{\theta_b} I + \varepsilon_b), \, \theta_b \sim p. 
    \vspace{-5pt}
    \label{eq:gen_loss}
\end{equation}
Notice that~\eqref{eq:gen_loss} is a differentiable function with respect to $I$. However, it involves sampling $\theta_b$ based on the distribution $p$, which is non-differentiable with respect to $p$. Thus, here the main question that we ask is: \textit{what is an alternative approximation for~\eqref{eq:gen_loss}, which is a differentiable function of $p$?}

  \begin{algorithm}[t]
   \caption{UVTomo-GAN}\label{alg:ctgan}
     \textbf{Require:} $\alpha_\phi$, $\alpha_I$, $\alpha_p$: learning rates for $\phi$, $I$ and $p$. $n_{\textrm{disc}}$: the number of iterations of the discriminator (critic) per generator iteration. $\gamma^I_{TV}$, $\gamma^I_{\ell_2}$, $\gamma^p_{TV}$, $\gamma^p_{\ell_2}$: the weights of total variation and $\ell_2$-regularizations for $I$ and $p$.\\
     \textbf{Require:} Initialize $I$ randomly and $p$ with $\textrm{Unif}(0, \pi)$.\\
     \textbf{Output:} Estimates $I$ and $p$ given $\{\xi^{\textrm{real}}_\ell\}_{\ell=1}^L$. 
   \begin{algorithmic}[1]
     \While{$\phi$ has not converged}
     \For{$t=0,...,n_{\textrm{disc}}$}
     \State Sample a batch from real data, $\{\xi^b_{\textrm{real}}\}_{b=1}^B$
     \State Sample a batch of simulated projections using estimated $I$ and $p$, i.e. $\{\xi^b_{\textrm{syn}}\}_{b=1}^B$ where $\xi^b_{\textrm{syn}} = \mathcal{P}_{\theta} I + \varepsilon_b$, $\varepsilon_b \sim \mathcal{N}(0, \sigma)$
     \State Generate interpolated samples $\{\xi^b_{\textrm{int}}\}_{b=1}^B$, $\xi^b_{\textrm{int}} = \alpha \, \xi^b_{\textrm{real}} + (1-\alpha) \, \xi^b_{\textrm{syn}}$ with $\alpha \sim \textrm{Unif}(0, 1)$
     \State Update the discriminator using gradient ascent steps using the gradient of~\eqref{eq:loss_function} with respect to $\phi$.
    \EndFor
     \State Sample a batch of $\{r_{i, b}\}_{b=1}^B$ using~\eqref{eq:gumbel_approx}
     \State Update $I$ and $p$ using gradient descent steps by taking the gradients of the following with respect to $I$ and $p$,
    \vspace{-5pt}
    {\tiny
     \begin{align}
         \mathcal{L}(I, p) &= \mathcal{L}_\mathcal{G}(I, p) + \gamma^I_{TV} \textrm{TV}(I) + \gamma^I_{\ell_2} \Vert I \Vert^2 + \gamma^p_{TV} \textrm{TV}(p) + \gamma^p_{\ell_2} \Vert p \Vert^2 \nonumber
     \end{align}}%
     \vspace{-20pt}
     \EndWhile
   \end{algorithmic}
 \end{algorithm}

To answer this question, we first discretize the support of the projection angles, i.e. $[0, \pi]$, uniformly into $N_{\theta}$ bins. Therefore, $p$ becomes a probability mass function (PMF), represented by a vector of length $N_\theta$ where $\sum\limits_{i=1}^{N_\theta} p_i = 1$, and $p_i \geq 0$, $\forall i$. This discretization has made the distribution over the projection angles discrete or categorical. In other words, the sampled projection angles from $p$ can only belong to $N_{\theta}$ discrete categories. This allows us to approximate~\eqref{eq:gen_loss} using the notions of gumbel-softmax distribution~\cite{gumbelsoftmax} as follows,  
\vspace{-5pt}
\begin{equation}
    \mathcal{L}_{\mathcal{G}}(I, p) \approx - \sum\limits_{b=1}^{B} \sum\limits_{i=1}^{N_\theta} r_{i, b} \mathcal{D}_{\phi} (\mathcal{P}_{\theta_i} I + \varepsilon_b), \vspace{-5pt} \label{eq:gen_aprox_loss}
\end{equation}
\vspace{-5pt}
with
\vspace{-5pt}
\begin{align}
    r_{i, b} \! = \! \frac{\exp{((g_{b,i} + \log(p_i))/\tau)}}{\sum\limits_{j=1}^{N_{\theta}} \exp{((g_{b,j} \! + \! \log(p_j))/\tau)}}, \, g_{b,i} \! \sim \! \textrm{Gumbel}(0, 1)
    \label{eq:gumbel_approx}
    \vspace{-5pt}
\end{align}
 where $\tau$ is the softmax temperature factor. As $\tau \rightarrow 0$, $r_{i, b} \! \rightarrow \! \textrm{one-hot} \left(\arg \max_i [g_{b,i} \! + \! \log(p_i)] \right)$. Furthermore, samples from the $\textrm{Gumbel}(0, 1)$ distribution are obtained by drawing $u \sim \textrm{Unif}(0, 1)$, $g \! = \! -\log(-\log(u))~\cite{gumbelsoftmax}$. Note that due to the reparametrization trick applied  in~\eqref{eq:gen_aprox_loss}, the approximated generator's loss has a tangible gradient with respect to $p$.
 
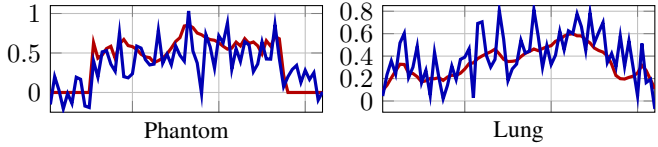
\begin{figure}
\centering
	\begin{tikzpicture}
		\begin{groupplot}[group style={group size= 2 by 2,                      
    				horizontal sep=0.8cm, vertical sep=0.1cm},     
	            	 legend pos= north east,        
					 legend style={legend cell align=left},
					 grid=both,                         
    				 height=3cm,width=5.2cm,
    				 xmin=1,xmax=64,
					 ymin=-0.25,ymax=1.1, 
					 ylabel near ticks, xlabel near ticks] 
		\nextgroupplot[x label style={at={(axis description cs:0.5,0)},anchor=north},xlabel={\small{Phantom}},xticklabels={,,}] 	
		\addplot[darkred,very thick] table[x=x, y=y, col sep=comma]{figs/viz_imgs/phantom_proj_clean_snr1.dat};
		\addplot[darkblue,very thick] table[x=x, y=y, col sep=comma]{figs/viz_imgs/phantom_proj_noisy_snr1.dat};
		
		\nextgroupplot[x label style={at={(axis description cs:0.5,0)},anchor=north},xlabel={\small{Lung}},xticklabels={,,}, ymin=-0.1,ymax=0.85] 	
		\addplot[darkred,very thick] table[x=x, y=y, col sep=comma]{figs/viz_imgs/body_proj_clean_snr1.dat};
		\addplot[darkblue,very thick] table[x=x, y=y, col sep=comma]{figs/viz_imgs/body_proj_noisy_snr1.dat};
		\end{groupplot}
	\end{tikzpicture}
\vspace{-1cm}
\caption{Examples of clean (red) and noisy (blue) projection lines for the experiments with $\textrm{SNR}=1$ in Fig.~\ref{fig:viz_results}.}
\vspace{-0.4cm}
\label{fig:proj_results}
\end{figure}
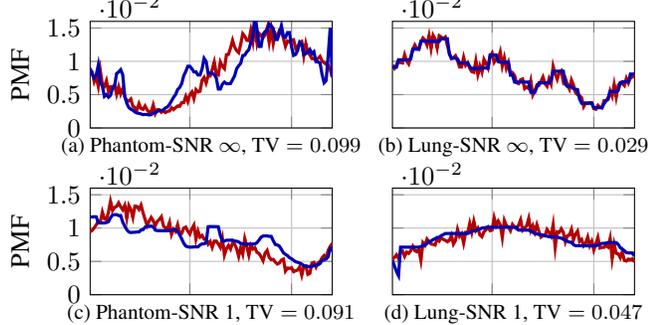
\begin{figure}
\centering
	\begin{tikzpicture}
		\begin{groupplot}[group style={group size= 2 by 2,                      
    				horizontal sep=0.8cm, vertical sep=0.8cm},     
	            	 legend pos= north east,        
					 legend style={legend cell align=left},
					 grid=both,                         
    				 height=3cm,width=4.8cm,
    				 xmin=1,xmax=120,
					 ymin=0,ymax=0.002, 
					 ylabel near ticks, xlabel near ticks] 
		\nextgroupplot[xticklabels={,,},ylabel={PMF},ymin=0,ymax=0.016,  x label style={at={(axis description cs:0.5,0)},anchor=north}, xlabel={\footnotesize{(a) Phantom-SNR $\infty$, $\textrm{TV}=0.099$}}] 	
			\addplot[darkred,very thick] table[x=x, y=y, col sep=comma]{figs/pdf_imgs/gt_pdf_phantom_nonoise.dat};
			\addplot[darkblue,very thick] table[x=x, y=y, col sep=comma]{figs/pdf_imgs/recon_pdf_phantom_nonoise.dat};

		\nextgroupplot[yticklabels={,,},xticklabels={,,},ymin=0,ymax=0.016, x label style={at={(axis description cs:0.5,0)},anchor=north}, xlabel={\footnotesize{(b) Lung-SNR $\infty$, $\textrm{TV}=0.029$}}] 
			\addplot[darkred,very thick] table[x=x, y=y, col sep=comma]{figs/pdf_imgs/gt_pdf_body_sigma0.dat};
			 \addplot[darkblue,very thick] table[x=x, y=y, col sep=comma]{figs/pdf_imgs/recon_pdf_body_sigma0.dat};
			 
		\nextgroupplot[xticklabels={,,},ylabel={PMF},ymin=0,ymax=0.016, xlabel={\footnotesize{(c) Phantom-SNR 1, $\textrm{TV}=0.091$}}, x label style={at={(axis description cs:0.5,0)},anchor=north}] 	
			\addplot[darkred,very thick] table[x=x, y=y, col sep=comma]{figs/pdf_imgs/gt_pdf_phantom_snr1.dat};
			\addplot[darkblue,very thick] table[x=x, y=y, col sep=comma]{figs/pdf_imgs/recon_pdf_phantom_snr1.dat};
			
		\nextgroupplot[xticklabels={,,},yticklabels={,,},ymin=0,ymax=0.016, x label style={at={(axis description cs:0.5,0)},anchor=north}, xlabel={\footnotesize{(d) Lung-SNR 1, $\textrm{TV}=0.047$}}] 	
			\addplot[darkred,very thick] table[x=x, y=y, col sep=comma]{figs/pdf_imgs/gt_pdf_body_snr1.dat};
			\addplot[darkblue,very thick] table[x=x, y=y, col sep=comma]{figs/pdf_imgs/recon_pdf_body_snr1.dat};
	
		\end{groupplot}
	\end{tikzpicture}
\vspace{-1cm}
\caption{Comparison between the ground truth sample distribution of the projection angles (red) and the one estimated by our method (blue). The setting of these experiments are the same as the ones in Fig.~\ref{fig:viz_results}.}
\vspace{-0.5cm}
\label{fig:pmf_results}
\end{figure} 
\input{figures/viz_imgs}

We present the pseudo-code for UVTomo-GAN in Alg.~\ref{alg:ctgan}. In all our experiments, we use a batch-size of $B=50$. 
We have three different learning rates for the discriminator, image and the PMF denoted by $\alpha_{\phi}$, $\alpha_I$ and $\alpha_{p}$. We reduce the learning rates by a factor of $0.9$, with different schedules for different learning rates. We use \texttt{SGD} as the optimizers for the discriminator and the image with a momentum of $0.9$ and update the PMF using gradient descent steps. We clip the gradients of the discriminator and the image by $1$ and $10$ respectively and normalize the gradients of the PMF. Following common practice, we train the discriminator $n_{\textrm{disc}}\!=\!4$ times per updates of $I$ and $p$. We discretize the domain of the projection angle, i.e. $[0, \pi]$, by roughly $2 d$ equal-sized bins, where $d$ is the image size. 

Due to the structure of the underlying images, we add $\ell_2$ and $\textrm{TV}$ regularization terms for the image, with $\gamma_{\ell_2}^I$ and $\gamma_{TV}^I$ weights. Furthermore, we assume that the unknown PMF is a piece-wise smooth function of projection angles (which is a valid assumption especially in single particle analysis in cryo-electron microscopy~\cite{punjani2017}), therefore adding $\ell_2$ and $\textrm{TV}$ regularization terms for the PMF with $\gamma_{\ell_2}^p$ and $\gamma_{TV}^p$ weights. 

Our default architecture of the discriminator consists of five fully connected (FC) layers with $2048$, $1024$, $512$, $256$ and $1$ output sizes. We choose ReLU~\cite{xu2015empirical} as the activation functions. To impose the non-negativity constraint over the image, we set $I$ to be the output of a $\texttt{ReLU}$ layer. In addition, to enforce the PMF to have non-negative values while summing up to one, we set it to be the output of a $\texttt{Softmax}$ layer. Our implementation is in PyTorch and we use Astra-toolbox~\cite{aarle2016astra} to define the tomographic projection operator.

\vspace{-0.25cm}
\section{Experimental Results}
\label{sec:results}
\vspace{-7pt}
We use two different images, a Shepp-Logan phantom and a biomedical image of lungs of size $64 \times 64$ in our experiments. We refer to these images as phantom and lung images throughout this section. We discretize the projection angle domain $[0, \pi]$ with $120$ equal-sized bins and generate a random piece-wise smooth $p$. We use this PMF to generate the projection dataset following \eqref{eq:proj_noisy}. We test our approach on a no noise regime (i.e. $\sigma=0$) and a noisy case where the signal-to-noise (SNR) ratio for the projection lines is $1$. For experiments with noisy phantom image, we use a smaller discriminator network with $512$, $256$, $128$, $64$ and $1$ as it leads to improved reconstruction compared to the default architecture. For all experiments the number of projection lines $L=20,000$. To assess the quality of reconstruction, we use peak signal to noise ratio (PSNR) and normalized cross correlation (CC). The higher the value of these metrics, the better the quality of the reconstruction. We use total variation distance ($\textrm{TV}$) to evaluate the quality of the recovered PMF compared to the ground truth. 

We compare the results of UVTomo-GAN with unknown PMF on four baselines, 1) UVTomo-GAN with known PMF, 2) UVTomo-GAN with unknown PMF but fixing it with a Uniform distribution during training, 3) TV regularized convex optimization, 4) expectation-maximization (EM). In the first baseline, similar to~\cite{cryogan}, we assume that the ground truth PMF of the projection angles is given. Thus, in Alg~\ref{alg:ctgan}, we no longer update $p$ (step 9). In the second baseline, we also do not update the PMF and during training assume that it is a Uniform distribution. In the third baseline, we assume that the angles associated to the projection lines are known, so formulate the reconstruction problem as a TV-regularized optimization solved using alternating direction method of multipliers (ADMM)~\cite{boyd2011distributed} and implement using GlobalBioIm~\cite{globalbioim}. In the fourth baseline, unlike the third one, we do not know the projection angles. Thus, we formulate the problem as a maximum-likelihood estimation and solve it via EM. 

\noindent \textbf{Quality of reconstructed image}: Figure~\ref{fig:viz_results} compares the results of UVTomo-GAN with unknown PMF against the ground truth image and the four baselines. Note that the results of UVTomo-GAN with unknown $p$ closely resembles UVTomo-GAN with known $p$, both qualitatively and quantitatively. However, with unknown $p$, the reconstruction problem is more challenging. Furthermore, we observe that with known $p$, UVTomo-GAN converges faster compared to the unknown $p$ case. Also, comparing the fourth and fifth columns in Fig.~\ref{fig:viz_results} shows the importance of updating $p$. While in the second baseline, the outline of the reconstructed images are reasonable, they lack accuracy in high-level details.

Note that while the first and third baselines are performing well on the reconstruction task, they have the advantage of knowing the projection angles or their distribution. Also, in our experiments we observed that EM is sensitive to the initialization. The EM results provided in Fig.~\ref{fig:viz_results} sixth column are initialized with low-pass filtered versions of the ground truth images. We observed that EM fails in successful detailed reconstruction if initialized poorly (Fig.~\ref{fig:viz_results} last column).

\noindent \textbf{Quality of reconstructed PMF}: Comparison between the ground truth distribution of the projection angles and the one recovered by UVTomo-GAN with unknown PMF is provided in Fig.~\ref{fig:pmf_results}. Note that the recovered PMF matches the ground truth distribution, thus proving the ability of our approach to recover $p$ under different distributions and noise regimes.
\vspace{-5pt}

\vspace{-0.25cm}
\section{Conclusion}
\vspace{-7pt}
In this paper, we proposed an adversarial learning approach for the tomographic reconstruction problem. We assumed neither the projection angles nor their probability distribution they are drawn from is known a-priori and we addressed the recovery of this unknown PMF alongside the image from the projection data. We formulated the reconstruction problem as a distribution matching problem which is solved via a min-max game between a discriminator and a generator. While updating the generator (i.e. the signal and the PMF), to enable gradient backpropagation through the sampling operator, we use gumbel-softmax approximation of samples from categorical distribution. Numerical results demonstrate the ability of our approach in accurate recovery of the image and the projection angle PMF. 



\vfill
\pagebreak
\newpage
\small{
\section{Compliance with Ethical Standards}
This is a numerical simulation study for which no ethical approval was required.}
\vspace{-10pt}
\small{
\section{Acknowledgement}
Mona Zehni and Zhizhen Zhao are partially supported by NSF DMS-1854791, NSF OAC-1934757, and Alfred P. Sloan Foundation.}
\vspace{-10pt}
\small{
\bibliographystyle{IEEEbib}
\bibliography{strings,refs}}

\end{document}